\documentclass[letterpaper, 10pt, conference]{ieeeconf}
\IEEEoverridecommandlockouts
\usepackage[pdftex]{graphicx}

\usepackage[font=footnotesize]{subcaption}
\usepackage[font=footnotesize]{caption}

\usepackage[nobreak]{cite}

\usepackage[ruled,linesnumbered,noend]{algorithm2e}
\SetAlCapFnt{\footnotesize}
\SetAlgorithmName{Alg.}{Alg.}{List of Algorithms}

\usepackage{cancel} 
\usepackage{times} 
\usepackage{amssymb, amsmath} 
\newcommand{\argmin}{\mathop{\rm arg~min}\limits} 

\usepackage{color}
\newcommand{\taharamod}{}

\usepackage{url}

\let\labelindent\relax 
\usepackage[inline]{enumitem}\setlist[enumerate]*{label=(\roman*)}
\usepackage[inline]{enumitem}\setlist[itemize]*{label=(\roman*)}

\newcommand\figref[1]{\text{Fig.~\ref{#1}}}
\newcommand\tableref[1]{\text{TABLE~\ref{#1}}}
\newcommand\equref[1]{\text{Eq.~(\ref{#1})}}
\newcommand\algorithmref[1]{\text{Alg.~\ref{#1}}}
\newcommand*{\sref}[1]{\S\ref{#1}}

\newcommand{\eg}{\textit{e.g.,}~}
\newcommand{\ie}{\textit{i.e.,}~}

\newcommand{\Naive}{Na\"ive }

\title{\LARGE \bf
    Disturbance Injection under Partial Automation: \\ Robust Imitation Learning for Long-horizon Tasks
}
\author{Hirotaka Tahara$^{1}$, Hikaru Sasaki$^{1}$, Hanbit Oh$^{1}$, Edgar Anarossi$^{1}$, and Takamitsu Matsubara$^{1}$
\thanks{$^{1}$The authors are with the Division of Information Science, Graduate School of Science and Technology, Nara Institute of Science and Technology (NAIST), Japan. This work is supported by JST [Moonshot Research and Development], Grant Number [JPMJMS2032].}
}

\begin{document}

\maketitle
\thispagestyle{empty}
\pagestyle{empty}

\begin{abstract}
Partial Automation (PA) with intelligent support systems has been introduced in industrial machinery and advanced automobiles to reduce the burden of long hours of human operation. Under PA, operators perform manual operations (providing actions) and operations that switch to automatic/manual mode (mode-switching). Since PA reduces the total duration of manual operation, these two action and mode-switching operations can be replicated by imitation learning with high sample efficiency. To this end, this paper proposes Disturbance Injection under Partial Automation (DIPA) as a novel imitation learning framework. In DIPA, mode and actions (in the manual mode) are assumed to be observables in each state and are used to learn both action and mode-switching policies. The above learning is robustified by injecting disturbances into the operator's actions to optimize the disturbance's level for minimizing the covariate shift under PA. We experimentally validated the effectiveness of our method for long-horizon tasks in two simulations and a real robot environment and confirmed that our method outperformed the previous methods and reduced the demonstration burden.
\end{abstract}

\section{INTRODUCTION}
\label{section_introduction}

\taharamod{\textit{Partial Automation} (PA) has been introduced in industrial machinery and advanced vehicles with intelligent assistance systems (\eg semi-automated driving cars\cite{huang2019data}, semi-automation of industrial machines such as construction\cite{rosenfeld1994enhancing, MorikawaIAARC2020}) to reduce the burden of long-time human operation\cite{feng2016synthesis}.} The operator performs manual operations (providing actions) and operations that switch to an automatic/manual mode (mode-switching). In the automatic mode, the operation is automated, although the operator monitors the task and switches to the manual mode when needed. In the manual mode, the operator manually performs the task and switches to the automatic mode whenever possible. By automating these two operations, a fully automated operation that resembles that of a skilled operator with PA is possible; this outcome is our paper's motivation.

\begin{figure}[tb]
\centering
\includegraphics[width=0.9\hsize] {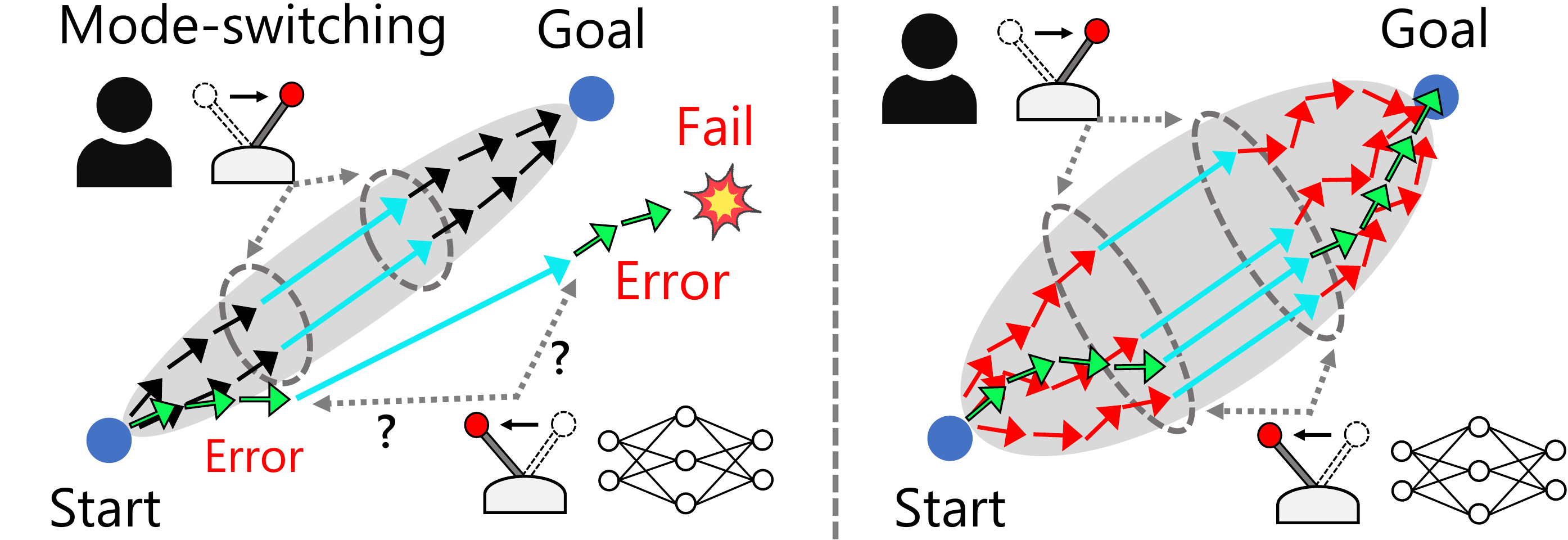}
    \begin{minipage}[b]{0.48\linewidth}
    \centering
    \subcaption{}
    \end{minipage}
    \begin{minipage}[b]{0.48\linewidth}
    \centering
    \subcaption{}
    \end{minipage}
\caption{
Reaching task by Imitation Learning (IL): (a) Operator demonstrates manual operation (black arrows) and mode-switching (arrow's color change) while using automatic operation (blue arrows) under Partial Automation (PA). A learned action policy (green arrows) and a mode-switching policy by \Naive IL under PA fail to complete the task due to the covariate shift that causes compounding errors. (b) In our method, disturbance injection into manual operation (red arrows) under PA expands the demonstrated distribution (gray area) and robustifies both action and mode-switching policies against covariate shift and accomplishes the task.
}
\label{fig:proopsed_method}
\end{figure}

This study aims to achieve such \textit{full automation} by Imitation Learning (IL) under PA for long-horizon tasks. IL is a promising approach for replicating the operator's behavior from demonstrations in a data-driven manner\cite{osa2018algorithmic, zhang2018deep, son2020expert}. Since PA reduces the total duration of manual operation, both action and mode-switching operations could be replicated by IL with high sample efficiency in terms of the number of demonstrations. However, a naive IL approach might not work due to the modeling errors of the learned policies that experience unknown states in the testing phase. This is called \textit{covariate shift} \cite{ross2010efficient}. Due to covariate shift, long-horizon prediction by learned policies creates compounding errors and fails to achieve tasks (\figref{fig:proopsed_method}(a)).

We address the issue by focusing on robust IL based on disturbance injection. A previous work achieved robust single policy learning by injecting disturbances into operator's actions to induce richer demonstrations\cite{laskey2017dart}. The disturbance's level was optimized based on the modeling error of the action policy to mitigate the covariate shift. Its direct application under PA is made by injecting disturbances into both actions and mode-switching; however, frequent mode changes increase the operator's cognitive load, which might lead to dangerous operation mistakes. \taharamod{Another approach would be to calculate disturbances based on the distribution extracted by demonstrated mode labels without the modeling error of the mode-switching policy and inject disturbances into actions only in manual mode; however, such disturbance might be insufficient due to underestimation of the covariate shift regarding mode-switching.}

\taharamod{We propose novel robust imitation learning framework called \textit{Disturbance Injection under Partial Automation} (\textbf{DIPA}) (\figref{fig:proopsed_method}(b)) for long-horizon tasks which can introduce PA.} In DIPA, both action and switching policies are learned using the operator's demonstrations that contain mode and actions, which are assumed to be observables in each state, collected by manual and mode-switching operations. \taharamod{Learning is robustified by injecting disturbances only into the manual operations, similar to a previous work \cite{laskey2017dart}; however, its disturbance's level is optimized for minimizing the \textit{mode-predictive covariate shift}. This novel objective function considers the modeling errors of the mode-switching policies by extracting the distribution using predicted modes.}

We conducted evaluations in two simulations (\textit{multi-object pick-and-place task} and \textit{soil-pile-flattening task}) and a real environment (\textit{soil-filling task}). In the simulations, the proposed method performed successfully on long-horizon tasks and outperformed the comparisons. The real experiment confirmed that our proposed method achieved the best performance while reducing the operator's demonstration burden.

\section{RELATED WORK}
\label{section_related_works}

\subsection{Automation of long-horizon tasks}
\label{section_related_work_long_horizon_tasks}

The challenging problem of fully automating long-horizon tasks using machine learning has been discussed \cite{mandlekar2020learning}. Hierarchical Reinforcement Learning (HRL) is one framework that focuses on policy decomposition. It divides a scheme into low-level policies for solving sub-tasks and high-level policies for selecting low-level policies \cite{barto2003recent, le2018hierarchical} in contrast to a single policy learning. This structure allows reward functions to be designed for each sub-task that prevents the sparsification of rewards and improves learning efficiency. However, this structure is problematic in terms of the cost of learning policies for each sub-task (e.g., reward design and the risk of failure due to trial-and-error explorations).

Our work focuses on Imitation Learning (IL) that does not require a reward function (although hybrid methods with RL and IL are beyond our focus\cite{le2018hierarchical, Li2021-tk}). Recent works \cite{Paradis2021IntermittentVS, Johns2021CoarsetoFineIL} proposed a coarse-to-fine IL that utilizes two types of policies: an open-loop, which does not consider the environment's state, and a closed-loop, which predicts the outputs considering the environment's state. This decomposition reduces the number of demonstrations and the subsequent demonstration burden in contrast to IL methods that learn a single policy. \taharamod{Unlike previous studies\cite{Paradis2021IntermittentVS,Johns2021CoarsetoFineIL} that utilize the task decomposition concept, our framework addressed the issue of compounding errors. In addition, we focus on the framework that considers policy learning problems for long-horizon tasks in contrast to one previous study \cite{Johns2021CoarsetoFineIL}, which focuses on IL for state estimation problems while utilizing some heuristically designed automatic controllers.}

\subsection{Robust Imitation Learning}
\label{section_related_work_robust_imitation_learning}

Policy learning by imitating demonstrator such as Behavior Cloning (BC) \cite{pomerleau1991efficient} is widely used in robot control \cite{zhang2018deep, osa2018algorithmic}. \taharamod{However, BC suffers from the compounding error during long-horizon prediction, where learned policies face unknown states that diverge from the observed distribution during training. This is known as \textit{covariate shift} \cite{ross2010efficient} that is induced by the modeling error of the learned policies.}

To mitigate covariate shift, such data augmentation methods as Dataset Aggregation (DAgger) \cite{ross2011reduction} and Disturbances for Augmenting Robot Trajectories (DART) \cite{laskey2017dart} have been proposed that robustify learned policies on augmented demonstrations. In \cite{ross2011reduction}, an expert collects recovery actions from compounding errors for executing the learned policy under the iterative training; however, this is potentially dangerous in interactions with the environment. Instead, we focus on methods with disturbance injection into a demonstrator's actions \cite{laskey2017dart, oh2021, taharaicra2022} to motivate collecting recovery actions from disturbances while iteratively optimizing disturbance's level based on the learned policy's modeling error.

However, the disturbance injection scheme \cite{laskey2017dart} assumes the robustification of a single policy. Under long-horizon tasks that require massive data for policy learning, the demand for recovery actions from disturbances increases the demonstration burden. Due to this assumption, demonstrators struggle to optimally perform tasks over a long period \cite{wu2019imitation}. Our work tackles this issue by employing the concept of policy decomposition to reduce the demonstration burden.

\section{PROPOSED METHOD}

\begin{figure}[t]
\centering
\includegraphics[width=0.9\hsize] {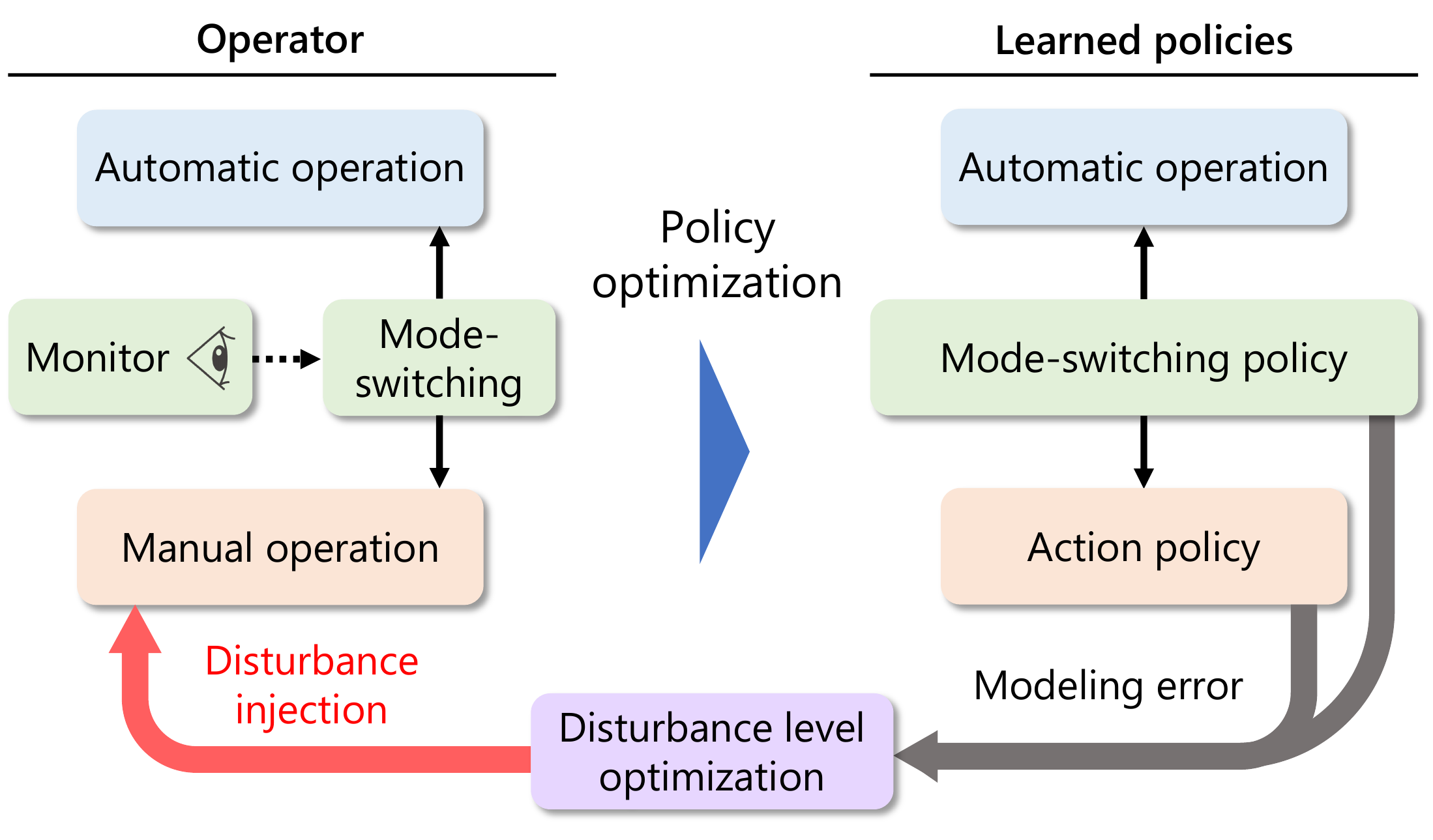}
\caption{ 
Overview of action and mode-switching policy robustification under partially automated demonstration in our framework DIPA
}
\label{fig:framework}
\end{figure}

An overview of our framework is shown in \figref{fig:framework}. Under Partial Automation (PA), an operator demonstrates mode-switching (+ monitoring) and manual operation while dealing with injected disturbances. If the operator selects the automatic mode, he focuses on monitoring the task during the automatic operation until mode-switching is required. This significantly reduces manual operation's total duration and demonstration load because the monitoring task's burden is less than the manual operation similar to industrial settings \cite{feng2016synthesis, rosenfeld1994enhancing, MorikawaIAARC2020, huang2019data}. After iterative policy and disturbance's level updating, the manual and mode-switching operation can be replaced with the learned policies utilizing automatic operation, and the system can become fully automated.

\subsection{Imitation Learning under Partial Automation }
\label{BCPA}

We formulate the basis of our method, BC \cite{pomerleau1991efficient} under Partial Automation (BCPA). BCPA aims to learn the action and mode-switching policies that replicate the operator's behavior under partially automated demonstrations. We define an action policy parameterized by $\theta^*$ as
\begin{align}
\pi_{\theta^*}(\mathbf{a}_t | \mathbf{s}_t, o_t) = 
\begin{cases}
\pi^{\mathrm{Manual}}_{\theta^*}(\mathbf{a}_t | \mathbf{s}_t) & \text{if } o_t = C\\
\pi^{\mathrm{Auto}}(\mathbf{a}_t | o_t) & \text{else } \\
\end{cases}, \label{eq:semi-automatic_policy}
\end{align}
and mode-switching policy as $\phi_{\lambda^*}(o_t | \mathbf{s}_t)$ parameterized by $\lambda^*$, where $*$ denotes the notations related to the operator. \taharamod{These operator's policies represent the operator's distribution.} $\pi^{\mathrm{Manual}}$ is the operator's manual action policy, and $\pi^{\mathrm{Auto}}$ is the automatic action policy that is predefined for each task. $\mathbf{s}_t$ is the N-dimensional state, $\mathbf{a}_t$ is the M-dimensional action, and a scalar value $o_t$ is the mode. $C \in \mathbb{N}$ is the index of $o_t$ for manual operation. We assume $o_t$ can be observed in the PA setting and collected through the operator's mode-switching operation. By using these policies, the trajectory distribution is defined:
\begin{align}
& p(\boldsymbol{\tau} | \lambda^*, \theta^*) \nonumber \\
& = p(\mathbf{s}_1) \prod_{t=1}^{T} \phi_{\lambda^*}(o_t | \mathbf{s}_t) \pi_{\theta^*}(\mathbf{a}_t | \mathbf{s}_t, o_t) p(\mathbf{s}_{t+1} | \mathbf{s}_t, \mathbf{a}_t), \label{eq:semi-automatic_trajectory_distribution}
\end{align}
where $\boldsymbol{\tau}=\left(\mathbf{s}_{1}, \mathbf{a}_1, o_1, \ldots, \mathbf{a}_{T-1}, o_{T-1}, \mathbf{s}_T\right)$ is a trajectory consisting of a series of states, actions, and modes, and $T$ is the total steps in a trajectory. $p(\mathbf{s}_{t+1} | \mathbf{s}_t, \mathbf{a}_t)$ is the environment's dynamics that follows Markov Decision Process.

To replicate the operator's behavior, the total loss between the operator's policies parameterized $\theta^*$ and $\lambda^*$ and query policies parameterized $\theta$ and $\lambda$ is defined:
\begin{align}
J(\lambda, \theta, \lambda^*, \theta^* | \boldsymbol{\tau}) = \sum_{t=1}^{T} & \big[ \mathcal{L}_\theta + \mathcal{L}_\phi \big],
\end{align}
where action loss $\mathcal{L}_\theta = \|\pi_{\theta}(\mathbf{s}_t, \phi_{\lambda}(\mathbf{s}_t)) - \pi_{\theta^*}(\mathbf{s}_t, \phi_{\lambda^*}(\mathbf{s}_t))\|_2^2$, and mode-switching loss $\mathcal{L}_\phi = |\phi_{\lambda}(\mathbf{s}_t) - \phi_{\lambda^*}(\mathbf{s}_t)|^2$. Policies $\pi_{\theta}(\mathbf{s}_t, \phi_{\lambda}(\mathbf{s}_t))$ and $\phi_{\lambda}(\mathbf{s}_t)$ are the deterministic form of the action and mode-switching policies. Since BCPA's objective is to minimize the expected loss along the trajectory distribution, we solve the following policy parameter optimization:
\begin{align}
\lambda^R, \theta^R = \argmin_{\lambda, \theta} \mathbb{E}_{ p(\boldsymbol{\tau} | \lambda^*, \theta^*) } J(\lambda, \theta, \lambda^*, \theta^* | \boldsymbol{\tau}),
\label{eq:semi-automatic_policy_update}
\end{align}
where $R$ denotes the notations related to the learned policies. Using these parameters, we define the learned action policy:
\begin{align}
\pi_{\theta^R}(\mathbf{a}_t | \mathbf{s}_t, o_t) = 
\begin{cases}
\pi^{\mathrm{Action}}_{\theta^R}(\mathbf{a}_t | \mathbf{s}_t) & \text{if } o_t = C\\
\pi^{\mathrm{Auto}}(\mathbf{a}_t | o_t) & \text{else } \\
\end{cases}, \label{eq:learned_semi-automatic_policy}
\end{align}
and the mode-switching policy as $\phi_{\lambda^R}(o_t | \mathbf{s}_t)$, where $\pi^{\mathrm{Action}}$ is the learned action policy to replace $\pi^{\mathrm{Manual}}$ in \equref{eq:semi-automatic_policy}.

However, BC-based IL policies suffer from compounding error caused by the covariate shift \cite{ross2010efficient}, which leads policies into undemonstrated states. The covariate shift, which is the difference in the trajectory distributions between the learned policies and the operator's policies, can be formalized:
\begin{align}
& \Big| \mathbb{E}_{ p(\boldsymbol{\tau} | \lambda^*, \theta^*) } J(\lambda^R, \theta^R, \lambda^*, \theta^* | \boldsymbol{\tau}) \nonumber \\ 
& - \mathbb{E}_{ p(\boldsymbol{\tau} | \lambda^R, \theta^R) } J(\lambda^R, \theta^R, \lambda^*, \theta^* | \boldsymbol{\tau}) \Big|.
\label{eq:semi-automatic_cov_shift}
\end{align}

\subsection{Robust Imitation Learning under Partial Automation}
\label{DIPA}

\begin{algorithm}[t]
\footnotesize
\SetAlgoLined
\DontPrintSemicolon
\SetKwInOut{Parameter}{Parameter}
\SetKwInOut{Initialize}{Initialize}
\SetKwInOut{Inputs}{Inputs}
\Parameter{Mode-switching policy $\phi_{\lambda^*}$, action policy $\pi_{\theta^*}$, disturbance parameter $\boldsymbol \Sigma$.}
\Initialize{Set the total number of iterations $K$, episodes $E$, and steps $T$. $s_1$ is an initial state. $\boldsymbol \Sigma_1$ is initialized with $\mathbf{0}$.}
\For{$k = 1$ to $K$}{
    \For{$e = 1$ to $E$}{
        \For{$t = 1$ to $T$}{
            Sample mode: $o_t \sim \phi_{\lambda^*}(o_t | \mathbf{s}_t)$
        
            Sample action: $\mathbf{a}_t \sim \pi_{\theta^*}(\mathbf{a}_t | \mathbf{s}_t, o_t, \boldsymbol \Sigma_{k})$ 
            
            Step environment: $\mathbf{s}_{t+1} \sim p(\mathbf{s}_{t+1} | \mathbf{s}_t, \mathbf{a}_t)$
            
        }
        Collected trajectories: $\boldsymbol{\tau}_{k}^e = \{ \mathbf{s}_{1:T}, \mathbf{a}_{1:T}, o_{1:T} \}$
    }
    Policy parameters $\lambda^R_{k+1}$ and $\theta^R_{k+1}$ are updated with Eq.~(\ref{eq:semi-automatic_disturbance_injected_policy_update}) \\
    Disturbance parameter $\boldsymbol \Sigma_{k+1}$ is updated with Eq.~(\ref{eq:semi-automatic_disturbance_update_analytical_solution})
}
\caption{\footnotesize Disturbance Injection under Partial Automation: DIPA}
\label{algorithm:Disturbance-injected semi-automatic Robust Imitation Learning}
\end{algorithm}

We formulate Disturbance Injection under Partial Automation (DIPA), which robustifies both action and mode-switching policies with the disturbance's level optimization on the mode-predictive covariate shift. \algorithmref{algorithm:Disturbance-injected semi-automatic Robust Imitation Learning} outlines DIPA.

\taharamod{
To address the issue of covariate shift, we focus on the disturbance injection approach\cite{laskey2017dart}. To induce richer demonstrations, \ie to simulate compounding errors under testing, disturbances $\boldsymbol{\epsilon}$ are injected into the operator's manual actions as, $\mathbf{a}_t = \pi^{\mathrm{Manual}}_{\theta^*}(\mathbf{s}_t) + \boldsymbol{\epsilon}_t$, where $\boldsymbol{\epsilon}_t$ sampled from the disturbance distribution is defined as $\boldsymbol{\epsilon}_t \sim p(\boldsymbol{\epsilon}\mid\psi)$, and the parameter $\psi$ denotes a sufficient statistic that defines the disturbance distribution. $\pi^{\mathrm{Manual}}_{\theta^*}(\mathbf{s}_t)$ is the deterministic form of the operator's manual action policy under disturbance injection $\pi^{\mathrm{Manual}}_{\theta^*}(\mathbf{a}_t | \mathbf{s}_t, \psi)$. Under the disturbances, the operator is forced to collect recovery actions that augment the demonstrated trajectory distribution to reduce the covariate shift of \equref{eq:semi-automatic_cov_shift}.} Based on \equref{eq:semi-automatic_policy}, the operator's action policy under the disturbance injection is defined:
\begin{align}
\pi_{\theta^*}(\mathbf{a}_t | \mathbf{s}_t, o_t, \psi) = 
\begin{cases}
\pi^{\mathrm{Manual}}_{\theta^*}(\mathbf{a}_t | \mathbf{s}_t, \psi) & \text{if } o_t = C\\
\pi^{\mathrm{Auto}}(\mathbf{a}_t | o_t) & \text{else } \\
\end{cases}, \label{eq:disturbance_injected_semi-automatic_policy}
\end{align}
where $o_t$ is selected by the operator's mode-switching policy $\phi_{\lambda^*}(o_t | \mathbf{s}_t)$. Importantly, disturbances are only injected into the manual action policy. Expanding the trajectory distribution by disturbance injection into the manual action policy simultaneously robustifies the mode-switching policy since the state space of the mode-switching policy is (partially) shared by the action policy.

By substituting the disturbance injected policies into \equref{eq:semi-automatic_trajectory_distribution}, the trajectory distribution with $\psi$ is defined:
\begin{align}
& p(\boldsymbol{\tau} | \lambda^*, \theta^*, \psi) \nonumber \\
& = p(\mathbf{s}_1) \prod_{t=1}^{T} \phi_{\lambda^*}(o_t | \mathbf{s}_t) \pi_{\theta^*}(\mathbf{a}_t | \mathbf{s}_t, o_t, \psi) p(\mathbf{s}_{t+1} | \mathbf{s}_t, \mathbf{a}_t).
\label{eq:disturbance_injected_semi-automatic_trajectory_distribution}
\end{align}
Since the covariate shift cannot be computed explicitly, the upper bound of \equref{eq:semi-automatic_cov_shift} is obtained by following DART \cite{laskey2017dart}:
\begin{align}
T \sqrt{\frac{1}{2} \mathrm{KL}\left( p(\boldsymbol{\tau} | \lambda^R, \theta^R) || p(\boldsymbol{\tau} | \lambda^*, \theta^*, \psi) \right) },
\end{align}
where $\mathrm{KL}(\cdot\| \cdot)$ is the KL divergence and can be expanded by canceling the common factors in the trajectory distributions:
\begin{align}
& \mathrm{KL}\left( p(\boldsymbol{\tau} | \lambda^R, \theta^R) || p(\boldsymbol{\tau} | \lambda^*, \theta^*, \psi) \right) \nonumber \\
& = \mathbb{E}_{ p(\boldsymbol{\tau} | \lambda^R, \theta^R) } \sum_{t=1}^{T} \log \frac{\phi_{\lambda^R}(o_t | \mathbf{s}_t) \pi_{\theta^R}(\mathbf{a}_t | \mathbf{s}_t, o_t)} {\phi_{\lambda^*}(o_t | \mathbf{s}_t) \pi_{\theta^*}(\mathbf{a}_t | \mathbf{s}_t, o_t, \psi)}.
\label{eq:KL_covariateshift}
\end{align}
To minimize the covariate shift, we optimize \equref{eq:KL_covariateshift} with respect to $\psi$ that makes the operator's distribution closer to the learned policies' distribution by solving below: 
\begin{align}
& \min_{\psi} \mathbb{E}_{ p(\boldsymbol{\tau} | \lambda^R, \theta^R) } - \sum_{t=1}^{T} \delta(o_t=C) \log \pi_{\theta^*}(\mathbf{a}_t | \mathbf{s}_t, o_t, \psi) \nonumber \\
& = \min_{\psi} \mathbb{E}_{ p(\boldsymbol{\tau} | \lambda^R, \theta^R) } L(\psi, \lambda^R, \theta^R, \theta^* | \boldsymbol{\tau}),
\label{eq:kl_minimization}
\end{align}
where manual operation is extracted by Dirac delta function $\delta(o_t=C)$ since disturbances are not injected into the automatic operation (\ie $\pi^{\mathrm{Auto}}$ in \equref{eq:disturbance_injected_semi-automatic_policy} does not have $\psi$).

\taharamod{\equref{eq:kl_minimization} minimizes the negative log-likelihood of the operator's action given the collected trajectories by the learned policies. However, the expectation cannot be solved since learned policy's trajectory distribution $p(\boldsymbol{\tau} | \lambda^R, \theta^R)$ is not observed before policy parameters $\lambda^R$ and $\theta^R$ have been trained. Therefore, we follow DART\cite{laskey2017dart} process that the operator's trajectory distribution with the $k$-th updated disturbance parameter $p(\boldsymbol{\tau} | \lambda^*, \theta^*, \psi_k)$ is used instead of $p(\boldsymbol{\tau} | \lambda^R, \theta^R)$ where $k$ is the number of optimization iterations. Also, the $\lambda^R$ and $\theta^R$ can be replaced with the $(k+1)$-th updated policy parameters $\lambda^R_{k+1}$ and $\theta^R_{k+1}$. The disturbance distribution for next iterations is updated to minimize the negative log-likelihood of the operator's action given the demonstrated trajectories by injecting disturbances with $\psi_k$: }
\taharamod{
\begin{align}
& \psi_{k+1} = \argmin_{\psi} \mathbb{E}_{ p(\boldsymbol{\tau} | \lambda^*, \theta^*, \psi_k) } L\left( \psi, \lambda^R_{k+1}, \theta^R_{k+1}, \theta^* | \boldsymbol{\tau} \right), \label{eq:semi-automatic_disturbance_update} \\
& L(\psi, \lambda^R_{k+1}, \theta^R_{k+1}, \theta^* | \boldsymbol{\tau}) \nonumber \\
& = - \sum_{t=1}^{T} \delta(o_t^R=C) \log \pi_{\theta^*} \left( \pi_{\theta^R_{k+1}}(\mathbf{s}_t, o_t^{R}) | \mathbf{s}_t, o_t^{R}, \psi \right) \label{eq:objective_semi-automatic_disturbance_update_with_delta_function} \\
& = - \sum_{t=1}^{T'} \log \pi^{\mathrm{Manual}}_{\theta^*} \left( \pi^{\mathrm{Action}}_{\theta^R_{k+1}}(\mathbf{s}_t) | \mathbf{s}_t, \psi \right),
\label{eq:objective_semi-automatic_disturbance_update} 
\end{align}
where the mode $o_t^R$ in \equref{eq:objective_semi-automatic_disturbance_update_with_delta_function} is obtained using the learned mode-switching policy $\phi_{\lambda^R_{k+1}}(\mathbf{s}_t)$, and $T'$ is the total time steps of manual operation extracted from $T$ using Dirac delta function $\delta(o_t^{R}=C)$. Here, we call \textit{Mode-predictive covariate shift} as the new concept of covariate shift that simultaneously minimizes the modeling error of the mode-switching policies along with the action policies.}

The policy parameters are updated using all the data until $k$ iterations:
\begin{align}
\lambda^R_{k+1}, \theta^R_{k+1} & = \argmin_{\lambda, \theta} \sum_{i=1}^{k} \mathbb{E}_{ p(\boldsymbol{\tau} | \lambda^*, \theta^*, \psi_i) } J(\lambda, \theta, \lambda^*, \theta^* | \boldsymbol{\tau}) \nonumber \\
& \approx \argmin_{\lambda, \theta} \frac{1}{E} \sum_{i=1}^{k} \sum_{e=1}^{E} J(\lambda, \theta, \lambda^*, \theta^* | \boldsymbol{\tau}_i^e),
\label{eq:semi-automatic_disturbance_injected_policy_update}
\end{align}
where $E$ is the total episodes (trajectories) and the sum of $E$ approximates the expectations on the trajectory distribution.

\subsection{DIPA with neural network model under Gaussian noise}

As an implementation of DIPA, deterministic deep neural network models are employed as learned action policy $\pi_{\theta^{\mathrm{nn}}}(\mathbf{s}_t, o_t)$ and mode-switching policy $\phi_{\lambda^{\mathrm{nn}}}(\mathbf{s}_t)$. \taharamod{In addition, we assume that the disturbance distribution follows Gaussian noise $\mathcal N(\mathbf{0}, \boldsymbol \Sigma)$ parameterized by variance $\boldsymbol \Sigma$. Under this assumption, the operator's manual actions are defined as $\mathbf{a}_t = \pi^{\mathrm{Manual}}_{\theta^*}(\mathbf{s}_t) + \boldsymbol{\epsilon}_t$ where $\boldsymbol{\epsilon}_t \sim \mathcal N(\mathbf{0}, \boldsymbol \Sigma)$. Then, the operator's manual action policy can be expressed as $\pi_{\theta^*}^{\mathrm{Manual}}(\mathbf{a}_t|\mathbf{s}_t,\psi) = \mathcal{N} \left(\mathbf{a}_t | \pi_{\theta^*}^{\mathrm{Manual}}(\mathbf{s}_t), \boldsymbol{\Sigma}\right)$ where $\psi$ is replaced with $\boldsymbol{\Sigma}$. According to the central limit theorem, it is reasonable to choose a Gaussian distribution for the disturbance distribution.} \taharamod{By substituting these components into \equref{eq:semi-automatic_disturbance_update} and \equref{eq:objective_semi-automatic_disturbance_update}, we obtained:
\begin{align}
\boldsymbol{\Sigma}_{k+1} = & \argmin_{\boldsymbol{\Sigma}} \mathbb{E}_{ p(\boldsymbol{\tau} | \lambda^*, \theta^*, \boldsymbol{\Sigma}_k) } \nonumber \\
& - \sum_{t=1}^{T'} \log \mathcal{N} \left( \pi^{\mathrm{Action}}_{\theta^{\mathrm{nn}}_{k+1}}(\mathbf{s}_t) | \pi^{\mathrm{Manual}}_{\theta^*}(\mathbf{s}_t), \boldsymbol{\Sigma} \right).
\label{eq:semi-automatic_disturbance_update_gaussian}
\end{align}
Since minimizing the negative log-likelihood under Gaussian distribution is equivalent to minimizing the quadratic function\cite{osa2018algorithmic}, the disturbance's level is optimized by Maximum Likelihood Estimation while approximating the expectation on the trajectory distribution:
\begin{align}
\boldsymbol{\Sigma}_{k+1} \approx \frac{1}{ET'} \sum_{e=1}^{E} \sum_{t=1}^{T'} \Big( \pi^{\mathrm{Action}}_{\theta^{\mathrm{nn}}_{k+1}}\big(\mathbf{s}_t \big) - \pi^{\mathrm{Manual}}_{\theta^*}(\mathbf{s}_t) \Big)^{2}.
\label{eq:semi-automatic_disturbance_update_analytical_solution}
\end{align}
}

The policy parameter optimization of \equref{eq:semi-automatic_disturbance_injected_policy_update} is solved by applying gradient descent methods. 

\begin{figure*}[tb]

    \centering
    \begin{minipage}[b]{0.27\linewidth}
    \centering
    \includegraphics[width=1.0\hsize]{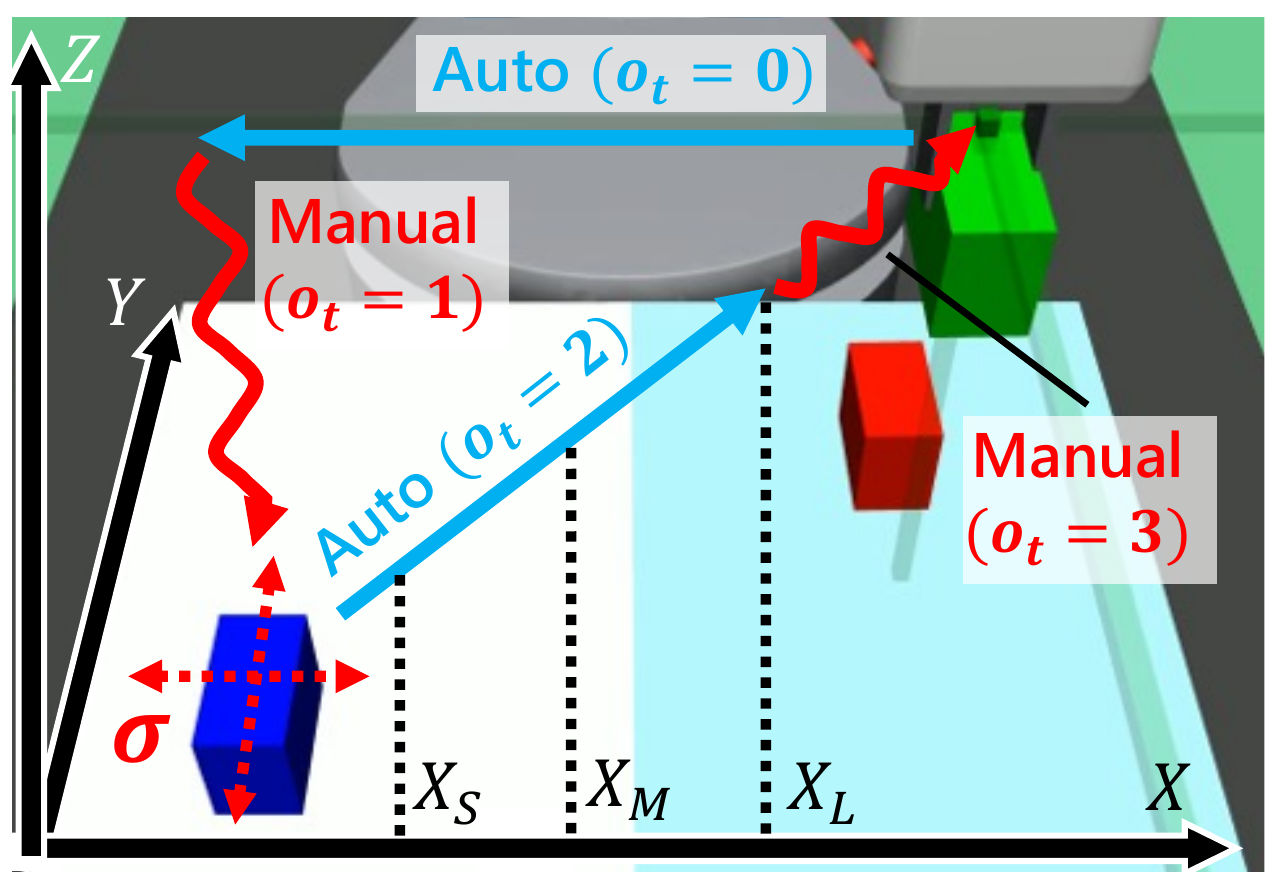}
    \subcaption{}
    \end{minipage}
    \begin{minipage}[b]{0.32\linewidth}
    \centering
    \includegraphics[width=1.0\hsize]{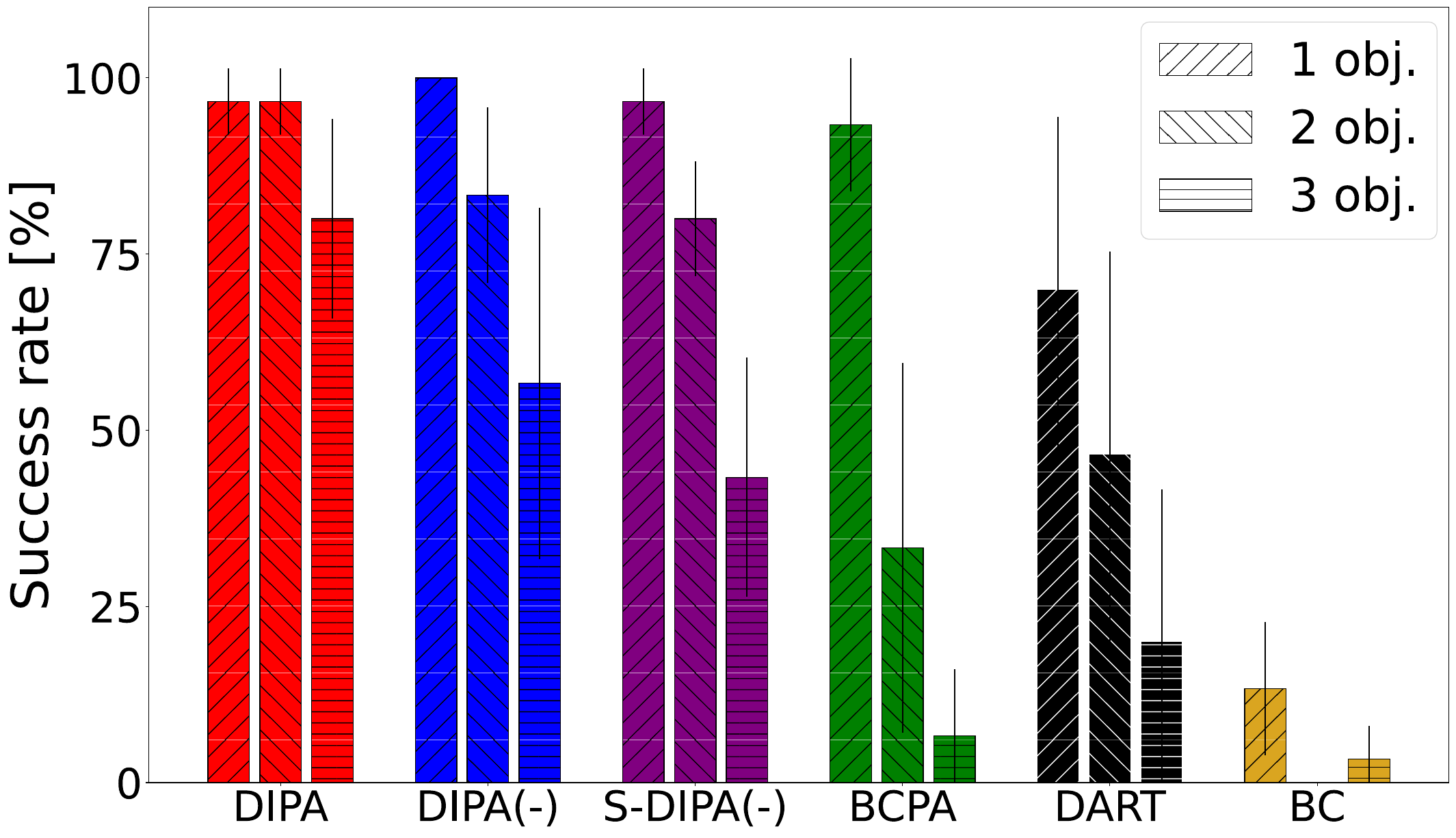}
    \subcaption{}
    \end{minipage}
    \begin{minipage}[b]{0.32\linewidth}
    \centering
    \includegraphics[width=1.0\hsize]{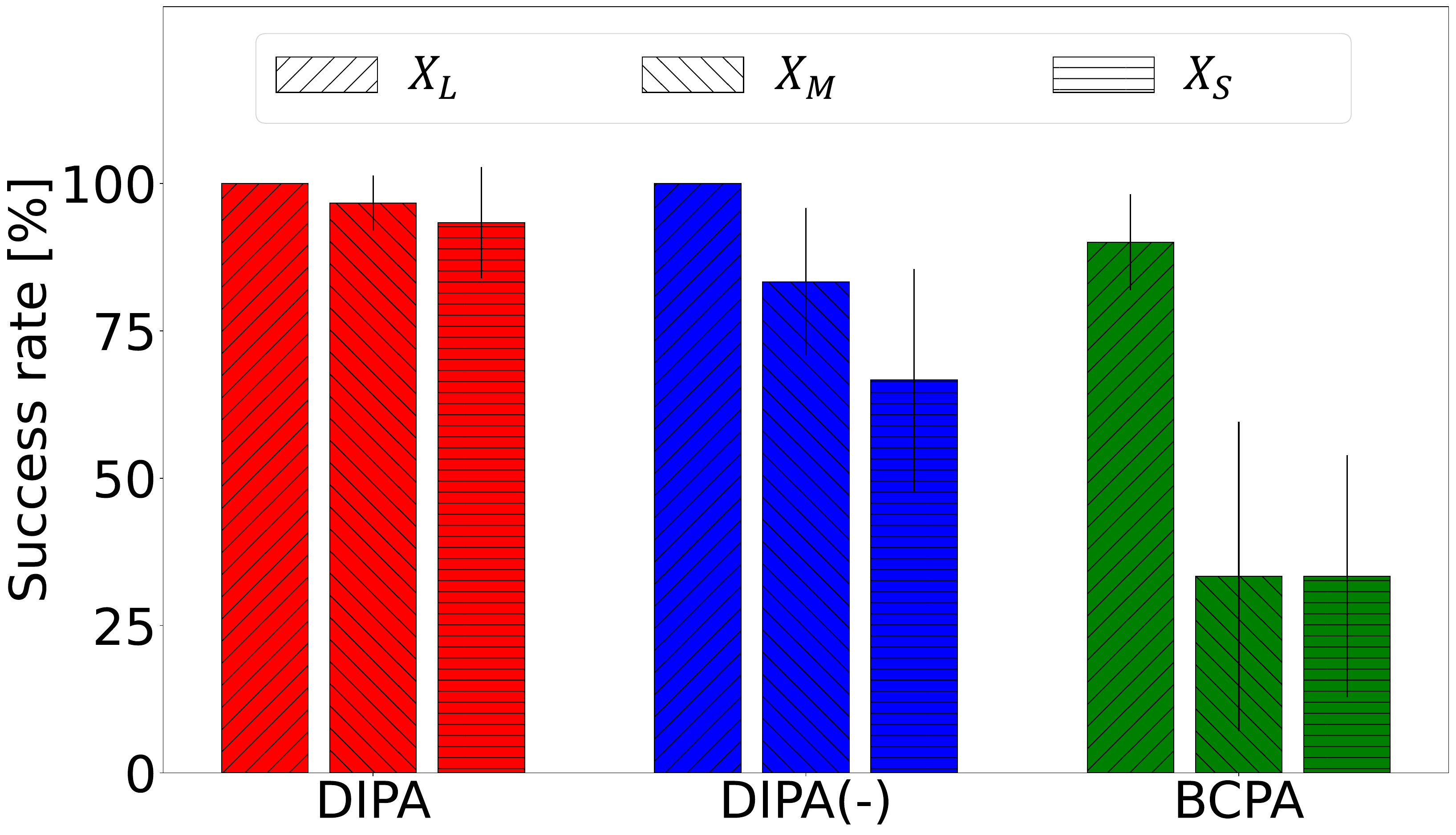}
    \subcaption{}
    \end{minipage}

    \centering
    \begin{minipage}[b]{0.33\linewidth}
    \centering
    \includegraphics[width=1.0\hsize]{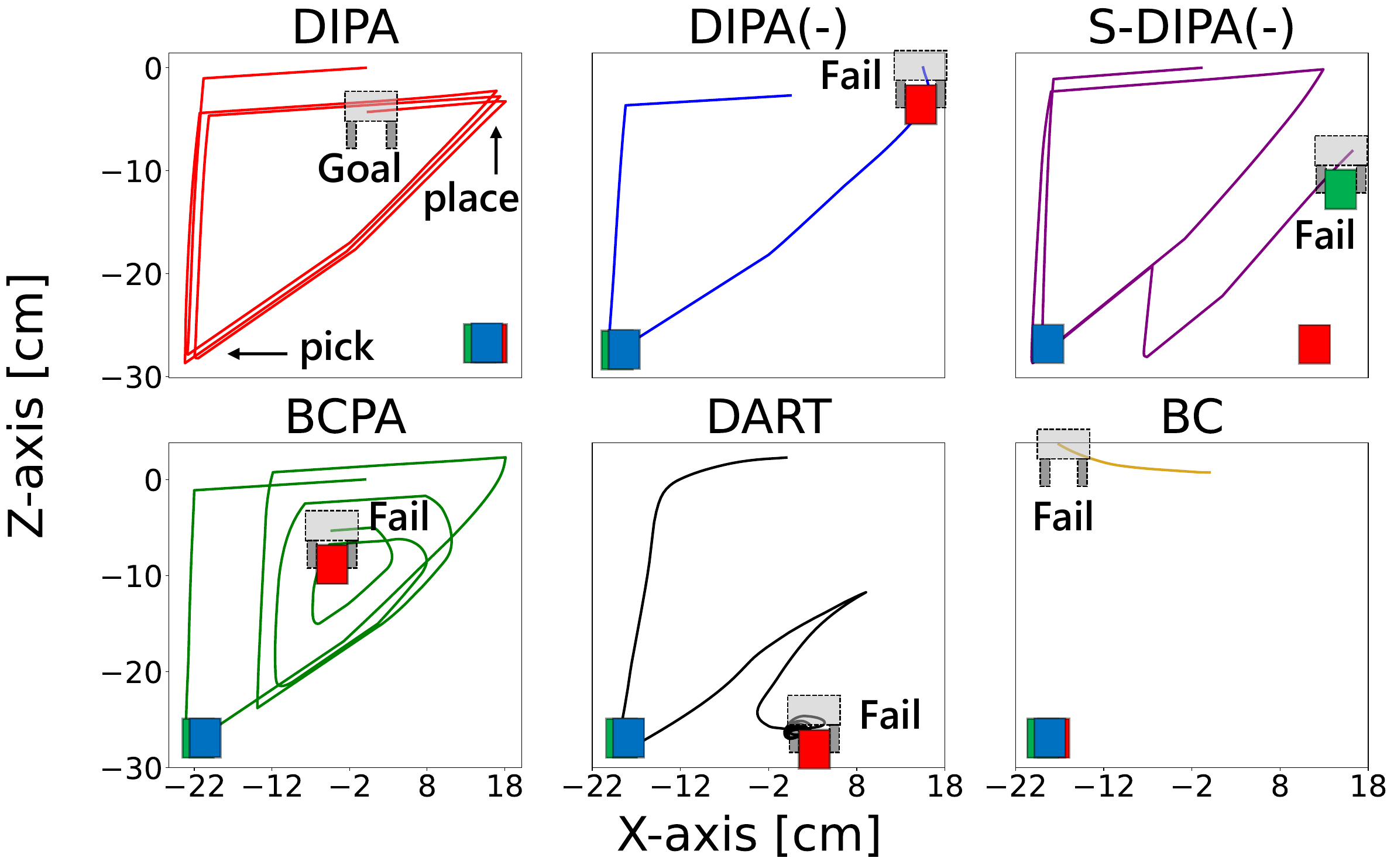}
    \subcaption{}
    \end{minipage}
    \begin{minipage}[b]{0.31\linewidth}
    \centering
    \includegraphics[width=1.0\hsize]{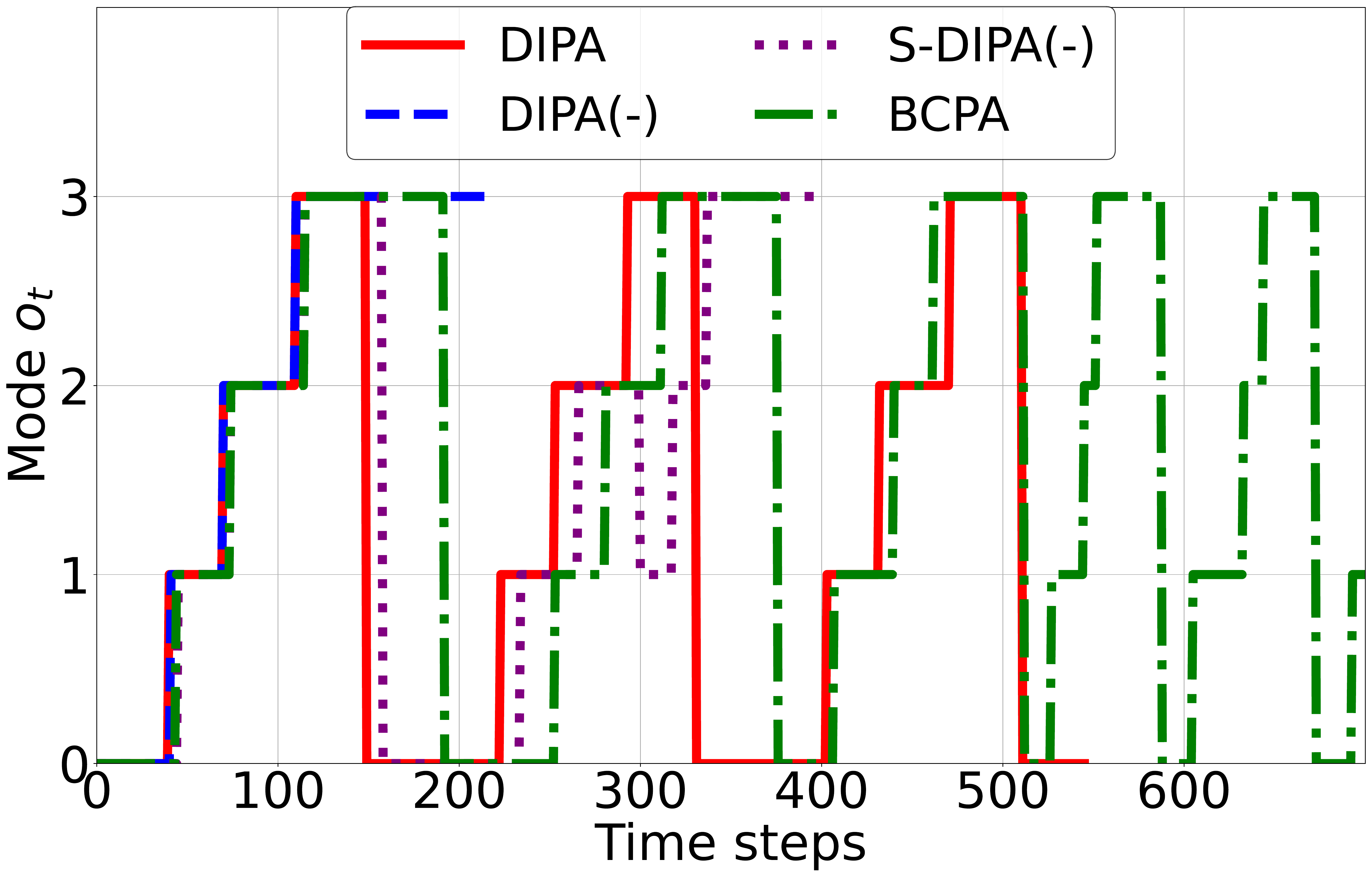}
    \subcaption{}
    \end{minipage}
    \begin{minipage}[b]{0.31\linewidth}
    \centering
    \includegraphics[width=1.0\hsize]{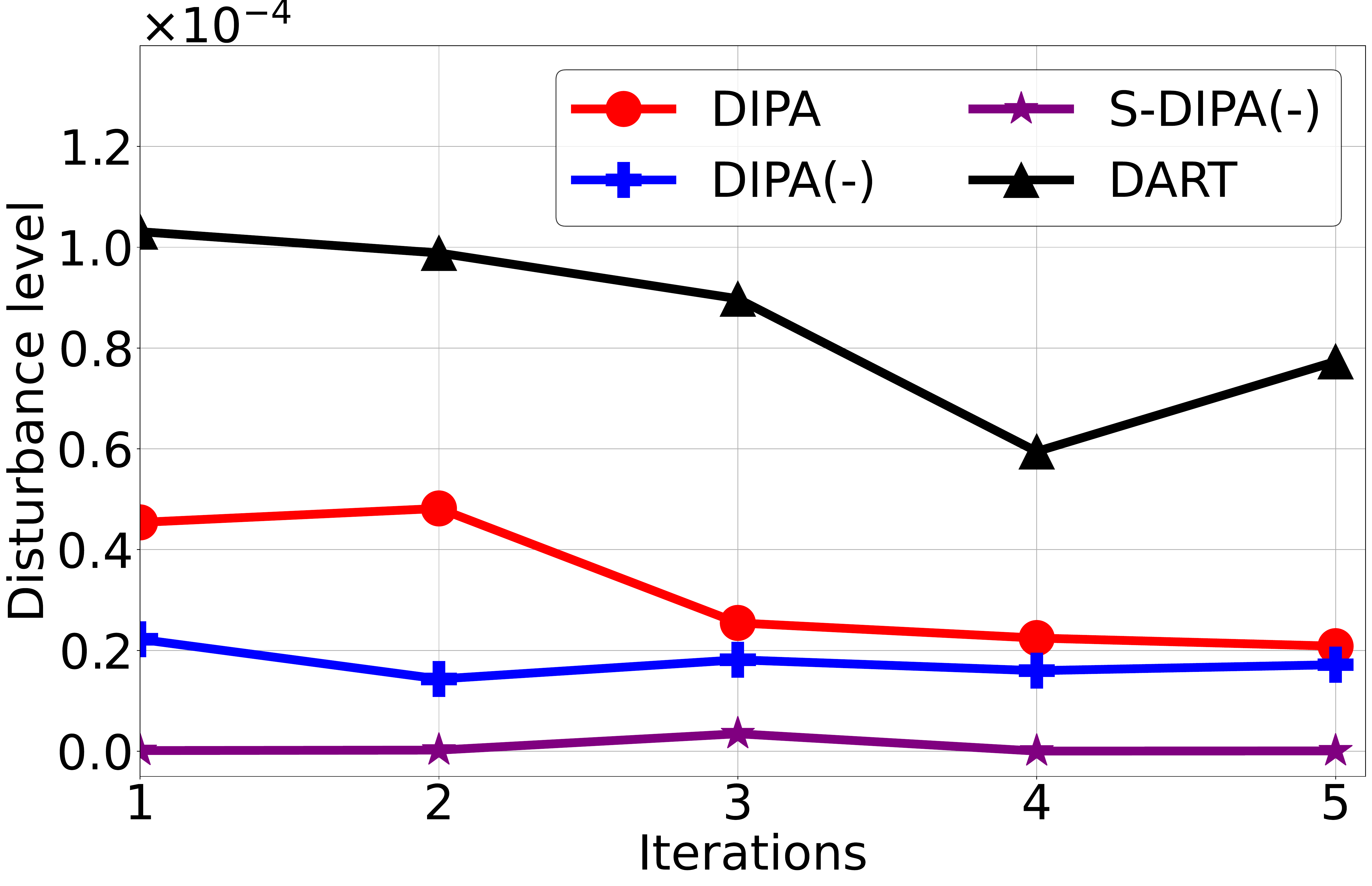}
    \subcaption{}
    \end{minipage}

\caption{
\taharamod{
(a) Multi-object pick-and-place task environment. 
(b) Test performances of one, two, and three objects. 
(c) Test performances under different designs of automatic action policies with two objects. 
(d) Test trajectories with three objects. 
(e) Modes along test trajectory in (d) predicted by learned mode-switching policy. 
(f) Result of disturbance's level updating.
}
}
\label{fig:experiment_pick_place}
\vspace{-3mm}
\end{figure*}

\begin{table}[t]
\caption{
\taharamod{Comparison methods. DIPA is our proposed method. DIPA(-), S-DIPA(-), and BCPA are the ablations. DART\cite{laskey2017dart} and BC\cite{pomerleau1991efficient} are the previous methods. Each method shows the availability of Partial Automation (PA) and disturbance injection (DI). DIPA, DIPA(-), and S-DIPA(-) also show the definition of the mode in Dirac delta function $\delta$ in \equref{eq:objective_semi-automatic_disturbance_update_with_delta_function}. Only S-DIPA(-) learns separated action policies for each manual mode.}
}
\label{table:ablations}
\centering
\begin{tabular}{c | c | c c c | c c}
\hline
            & DIPA       & DIPA(-)       & S-DIPA(-)     & BCPA          & DART         & BC \\
\hline
PA          & $\checkmark$  & $\checkmark$  & $\checkmark$  & $\checkmark$  & --            & -- \\
DI          & $\checkmark$  & $\checkmark$  & $\checkmark$  & --            & $\checkmark$  & -- \\
$\delta$ & $o_t^R $ & $o_t^*$ & $o_t^*$ &N.A. & N.A. & N.A. \\
\hline
\end{tabular}
\end{table}

\section{Simulation}
\label{section_simulation}

This section evaluates the effectiveness of our method for long-horizon tasks. We have two aims (P1, P2) for the validation:
\begin{itemize*}
\item[(P1)] compares the robustness of comparisons under multiple task settings, and 
\item[(P2)] investigates its robustness under different designs of automatic action policies.
\end{itemize*}
(P1, P2) are validated on two challenging long-horizon tasks \cite{Li2021-tk, dadhich2016key}:
\begin{enumerate*}
\item \textit{multi-object pick-and-place task} to validate (P1, P2) and 
\item \textit{soil-pile-flattening task} to validate (P1).
\end{enumerate*}

\taharamod{In both simulations, we employ a heuristically designed PID-based algorithmic operator instead of a human operator to keep the demonstration quality. This setting intends that experiments fairly evaluate the accuracy of the learned policies rather than the demonstration burden that causes performance degradation issues. In contrast, the latter user study evaluation, which is harder to prepare an algorithmic operator by hand, employs human operators.}

\taharamod{\tableref{table:ablations} shows the comparison methods. DIPA is our method explained in \sref{DIPA}. DIPA(-) and Separated-DIPA(-); S-DIPA(-) for short, are the ablations of DIPA which replace $o_t^R$ in \equref{eq:objective_semi-automatic_disturbance_update_with_delta_function} with the demonstrated mode $o_t^*$ provided by the operator's mode-switching policy $\phi_{\lambda^*}(\mathbf{s}_t)$. Therefore, DIPA(-) and S-DIPA(-) ignore the modeling error of the mode-switching policy in disturbance calculation as mentioned in \sref{section_introduction}. S-DIPA(-) validated in \sref{section_pick_place} learns separated action policies for each manual mode, although other methods learn a combined action policy for all manual modes. BCPA has already been explained in \sref{BCPA}. DART\cite{laskey2017dart} and BC\cite{pomerleau1991efficient} are previous methods without PA explained in \sref{section_related_work_robust_imitation_learning}.}

\subsection{Multi-object pick-and-place simulation}
\label{section_pick_place}

\subsubsection{Task setting}

\figref{fig:experiment_pick_place}(a) shows the customized OpenAI Gym environment with a robot arm. The task goal is to move every object from the white to the blue area. The number of objects ($N_{obj.}$) increases from one to three, which lengthens the task. To complicate the task, we added uniform noise $\mathcal{U}(-\sigma, \sigma)$ where $\sigma = 2~\mathrm{cm}$ in the initial positions ($X$ and $Y$) of the objects when the environment is initialized.


\subsubsection{Partial Automation setting}

\taharamod{In our evaluations, the tasks are designed following two guidelines:
\begin{enumerate*}
\item for introducing PA, a single long-horizon task is decomposed into a minimum number of sub-tasks based on expert knowledge to reduce the mode-switching demonstration burden\cite{hoque2021thriftydagger}, and
\item each sub-task is classified as manual or automatic operations. The manual operation is used for the state-dependent operation requiring human decisions, and the others, where less interaction with the environment, are replaced by automatic operations\cite{Paradis2021IntermittentVS}.
\end{enumerate*}}
Following guidelines, we decompose the robot motion into four modes (\figref{fig:experiment_pick_place}(a)): moving the gripper to the white area in the automatic mode ($o_t=0$); reaching the object at a random initial position in the manual mode ($o_t=1$); moving the gripper to the blue area in the automatic mode ($o_t=2$); placing the object in a certain position in the manual mode ($o_t=3$). \taharamod{For a partially automated demonstration, the mode transition is limited as $o_t:0\rightarrow1\rightarrow2\rightarrow3\rightarrow0$ to reduce the cognitive load associated with mode-switching, and the operator manually changes to the next mode at the end of each mode.}

\taharamod{To verify the aim (P2), we applied the automatic action policy ($o_t = 2$) with the three different designs of thresholds, which represent the mode-switching positions by the algorithmic operator. Thresholds are shown as dotted lines in Fig. 3(a) where $X_L$, $X_M$, and $X_S$ are defined as $+15~\mathrm{cm}$, $0~\mathrm{cm}$, and $-15~\mathrm{cm}$, respectively. The automatic mode ($o_t = 2$) is switched to the following manual mode ($o_t = 3$) when the gripper $X$-axis position exceeds each threshold.}

\begin{figure*}[tb]

    \begin{minipage}[b]{0.31\linewidth}
    \centering
    \includegraphics[width=1.0\hsize]{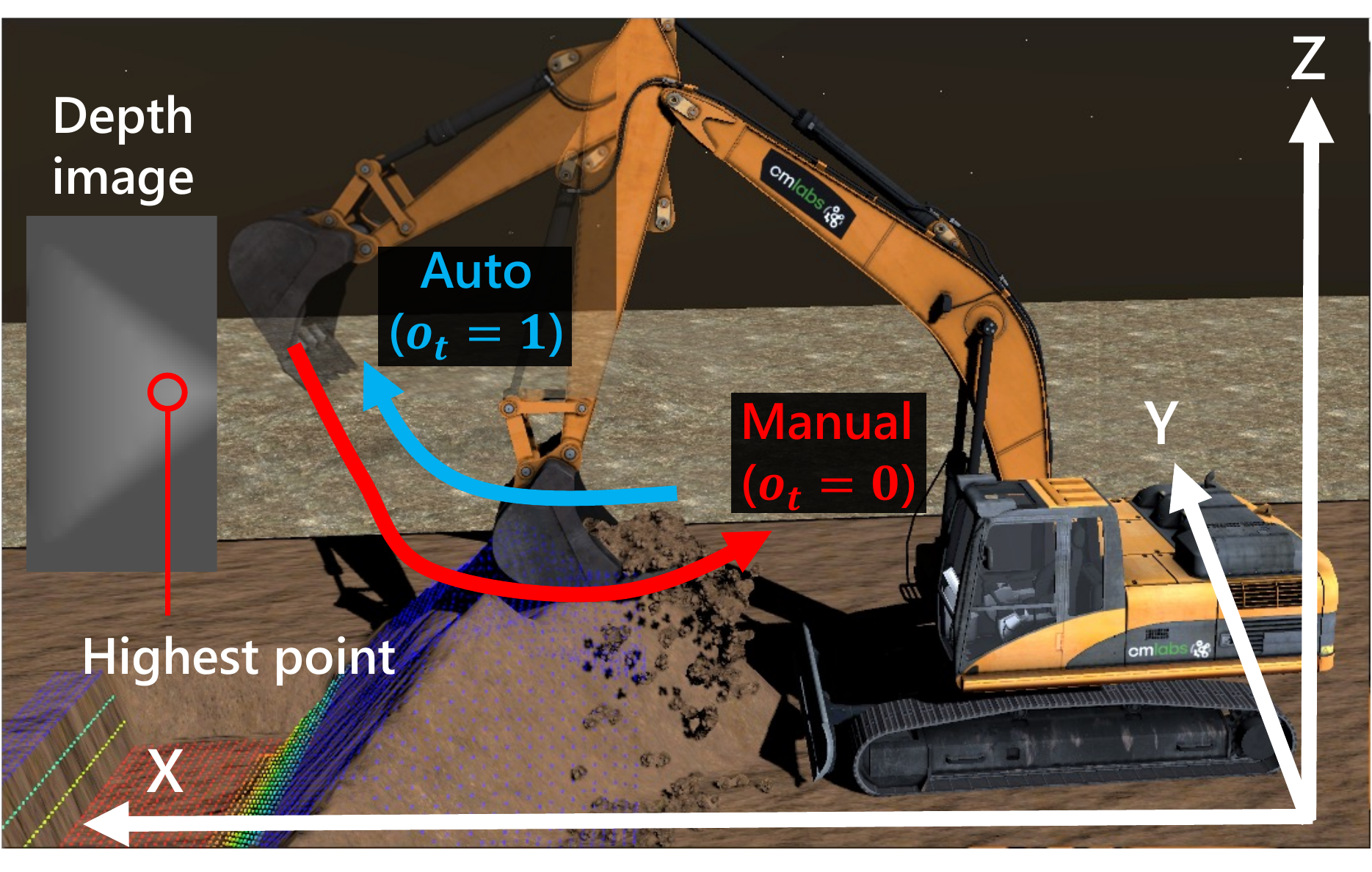}
    \subcaption{}
    \end{minipage}
    \begin{minipage}[b]{0.34\linewidth}
    \centering
    \includegraphics[width=1.0\hsize]{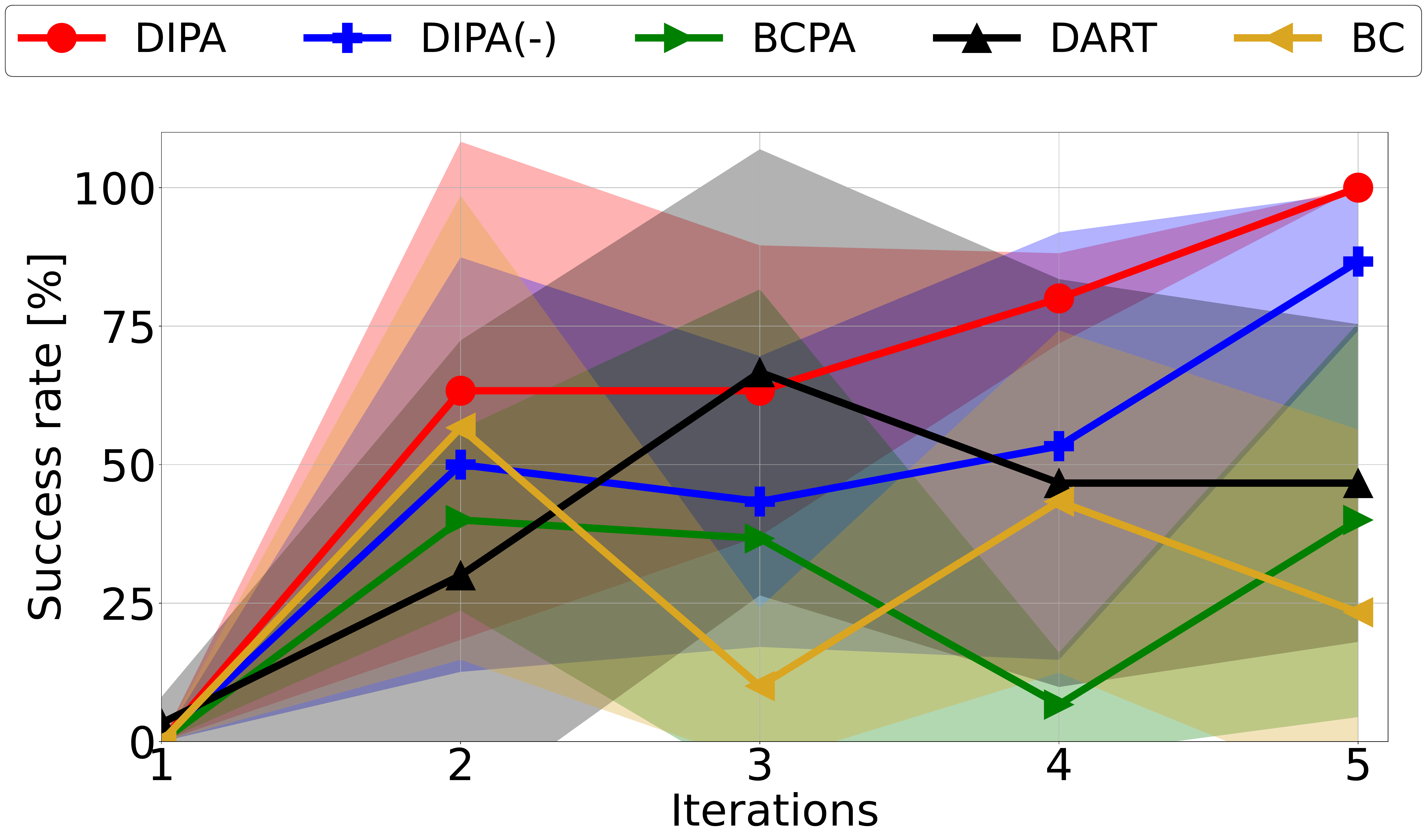}
    \subcaption{}
    \end{minipage}
    \begin{minipage}[b]{0.31\linewidth}
    \centering
    \includegraphics[width=1.0\hsize]{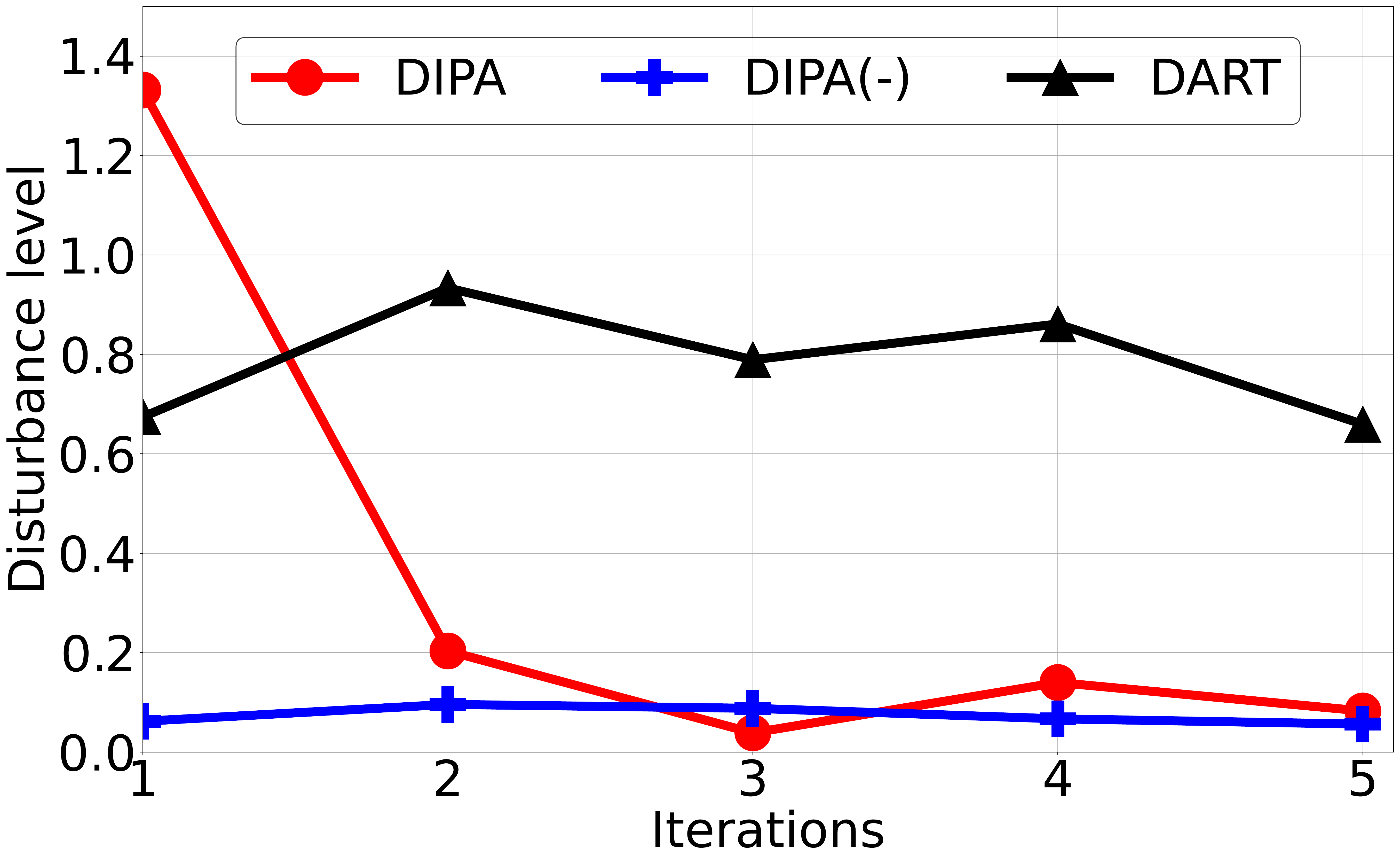}
    \subcaption{}
    \end{minipage}
    
\caption{(a) Soil-pile-flattening task environment. (b) Iterative test performances (mean: solid lines, STD: shaded). (c) Result of disturbance's level updating.}
\label{fig:experiment_vortex}
\vspace{-3mm}
\end{figure*}


\subsubsection{Demonstration setting}

The states and actions are defined as follows: 
\begin{itemize*}
\item[\textbf{(i) action policy w/o PA:}] the states are the gripper position and joint angle $(X,Y,Z,\theta)$, the object positions $(\{X,Y,Z\} \times N_{obj.})$, and the number of moved objects. \taharamod{The actions are the deviation of gripper position and joint angle at each step $(\Delta{X},\Delta{Y},\Delta{Z},\Delta{\theta})$.}
\item[\textbf{(ii) action policy with PA:}] \taharamod{the states are the gripper position, object positions, and the number of moved objects without the gripper joint angle $\theta$ since the automatic policy controls the grasping}, and the actions are identical as (i). 
\item[\textbf{(iii) mode-switching policy with PA:}] the states are identical as (i), and the action is the mode. 
\end{itemize*}
Automatic action policies $\pi^{\mathrm{Auto}}(o_t)$ are defined as follows: if $o_t = 0$, moving the gripper to the left while opening the gripper \taharamod{$(\Delta{X},\Delta{Y},\Delta{Z},\Delta{\theta}) = (-5~\mathrm{cm}, 0~\mathrm{cm}, 0~\mathrm{cm}, 1~\mathrm{rad})$}, and if $o_t = 2$, moving the gripper to the upper right while closing the gripper \taharamod{$(\Delta{X},\Delta{Y},\Delta{Z},\Delta{\theta}) = (5~\mathrm{cm}, 0~\mathrm{cm}, 3~\mathrm{cm}, -1~\mathrm{rad})$.}

\subsubsection{Learning setting}

We collected demonstrations for five iterations with ten episodes each (100 to 500 steps per episode depending on the number of objects). The policies were trained with three seeds over 300 epochs using early stopping. The policy model employs DART architecture \cite{laskey2017dart}, a four-layer deep neural network consisting of an input layer, two 64-unit hidden layers, and an output layer. The learned policies are evaluated ten times each in the testing phase.

\subsubsection{ (P1) result}

\taharamod{
\figref{fig:experiment_pick_place}(b) shows the test performances of all methods. DIPA, DIPA(-), S-DIPA(-), and BCPA achieved high performance for one object, in contrast to BC, which does not have PA or disturbance injection, resulting unstably. This result indicates the importance of policy decomposition and policy robustification. Although DART performs better than BC, DART's performance degradation becomes more apparent when the task is lengthened due to the increased number of objects. These results indicate that BC and DART, which learn a single policy for combined modes, are inferior to our DIPA, which remains consistently stable. In addition, \figref{fig:experiment_pick_place}(d) and (e) show the comparison methods (DIPA(-), S-DIPA(-), and BCPA) failed due to the propagated compounding errors through the mode-switchings that caused the fatal failure mode-switchings. In contrast, DIPA robustly achieved the task with the significant robustness of the mode-switching policies even though the number of mode-switchings increased.
}

\taharamod{
\figref{fig:experiment_pick_place}(f) shows the updates of the disturbance's level. DART suffered from significant modeling errors due to the difficulty of learning the long-horizon task. The error caused a prominent disturbance's level with slow convergence that potentially induced an unexpectedly broad operator's demonstrations, reducing data density. This prevented decreasing the modeling error and resulted in DART's low performance. In contrast, our DIPA that optimizes the disturbance's level to minimize the mode-predictive covariate shift resulted in reasonable disturbances. On the other hand, DIPA(-) and S-DIPA(-), which ignore the modeling error of the mode-switching policy, underestimated the disturbance's level, causing premature convergence. This resulted in DIPA(-), and S-DIPA(-) being unable to robustify the mode-switching policy sufficiently and caused the performance shown in \figref{fig:experiment_pick_place}(b). In addition, S-DIPA(-), which learns separated action policies, resulted in a further smaller disturbance's level than DIPA(-), resulting in less robustness. 
}

\subsubsection{(P2) result}
\taharamod{
\figref{fig:experiment_pick_place}(c) shows the test performances with three different designs of thresholds $X_L$, $X_M$, $X_S$ for automatic action policy under two objects. The $X_L$ did not make performance differences, similar to \cite{Paradis2021IntermittentVS}. However, the performance of BCPA significantly deteriorated with $X_M$ and $X_S$, which lengthened the learned policy's work. This indicated that the design of automatic action policies affects the test performances. In contrast, DIPA and DIPA(-) experienced less performance degradation owing to policy robustification by disturbance injection.
}

\subsection{Soil-pile-flattening simulation}
\label{section_vortex}

\subsubsection{Task setting}

\figref{fig:experiment_vortex}(a) shows the CM Labs VortexStudio environment, which enables real-time soil simulation. The task goal is to level the soil pile at a certain height. A depth camera is installed to capture the soil shape from the top of the environment, and the image is updated when the robot is not in the image (no occlusion). Uniform noise $\mathcal{U}(-\sigma, \sigma)$ with $\sigma=5~\%$ is added to the initial soil density ($65~\%$) that corresponds to the soil hardness when the environment is initialized. Hard soil pile requires more excavation to break it, while soft requires less excavation.

\begin{figure*}[tb]
    \centering
    \begin{minipage}[c]{0.66\linewidth}
    \centering
    \includegraphics[width=1.0\hsize]{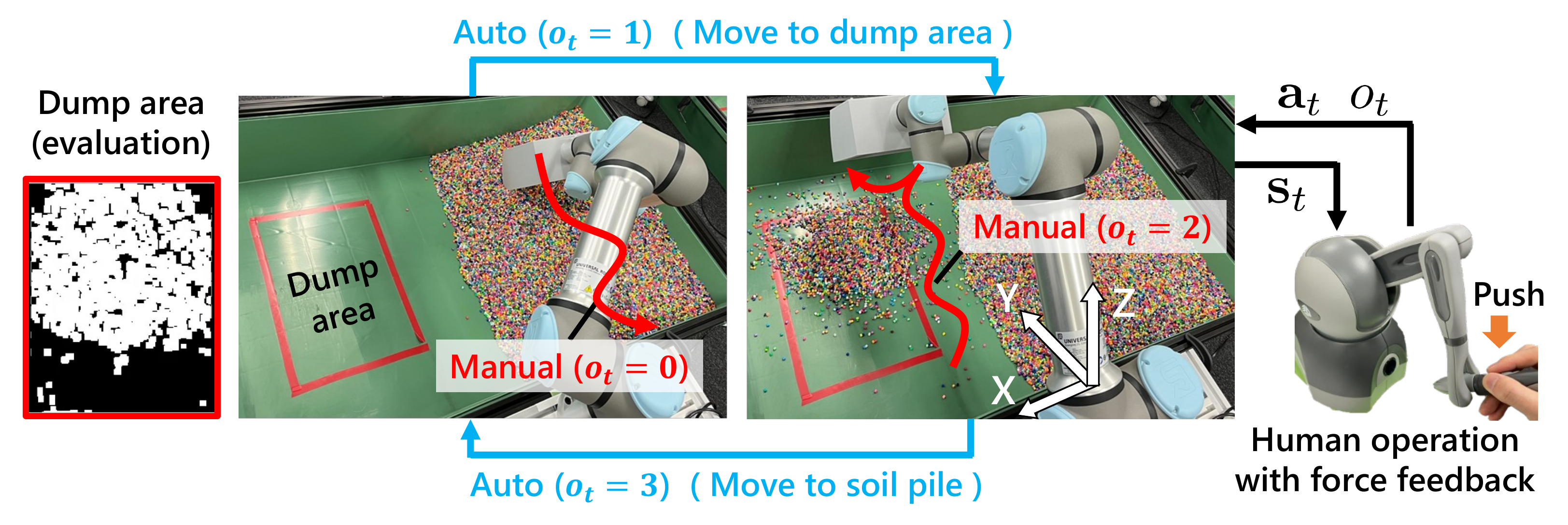}
    \subcaption{}
    \end{minipage}
    \begin{minipage}[c]{0.28\linewidth}
    \centering
    \includegraphics[width=1.0\hsize]{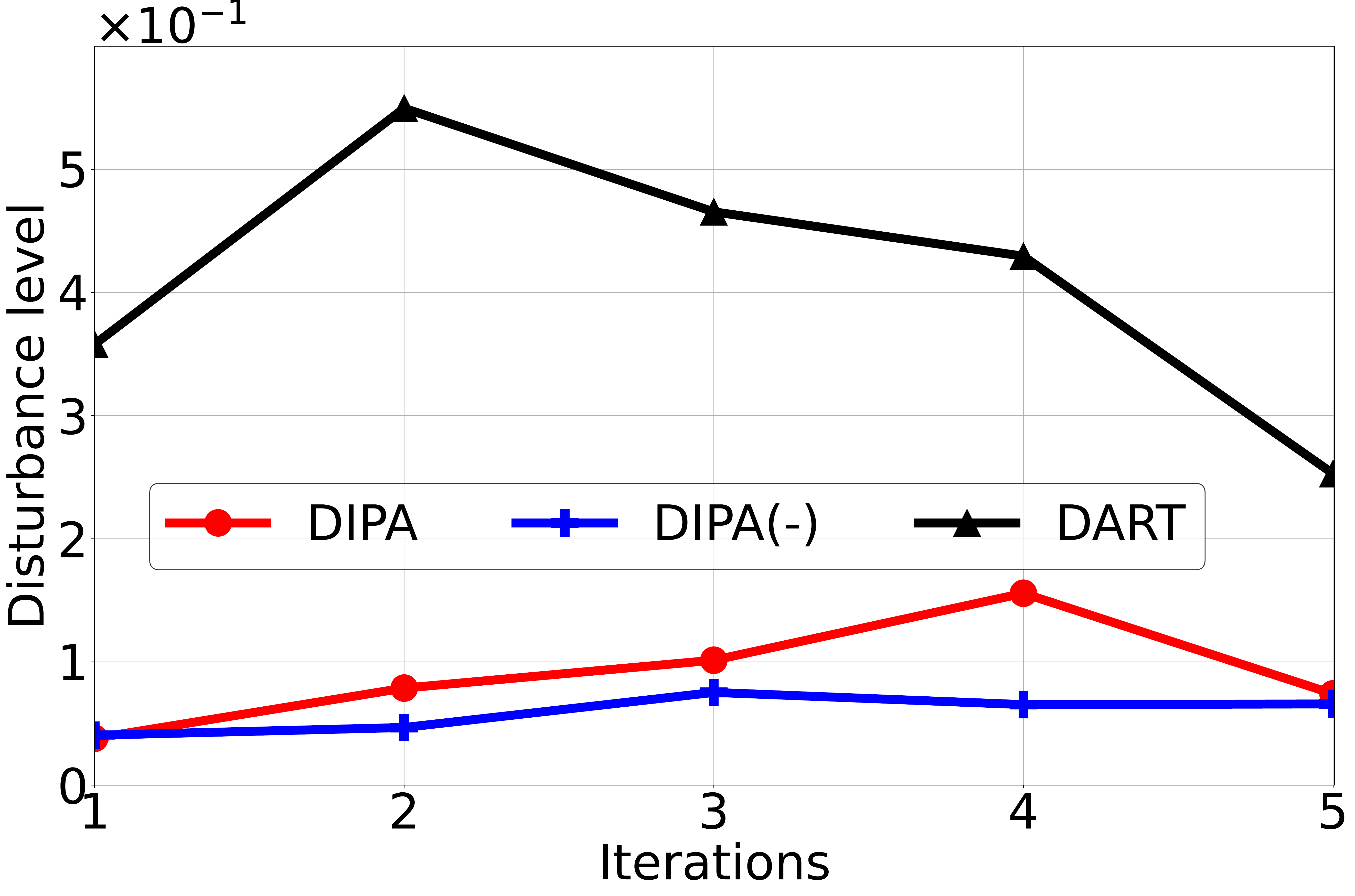}
    \subcaption{}
    \end{minipage}
    
\caption{(a) Real robot environment for soil-filling task with a robot arm under human demonstrations. (b) Result of disturbance's level updating.}
\label{fig:experiment_real_robot}
\vspace{-3mm}
\end{figure*}

\subsubsection{Partial Automation setting}

We decomposed the robot motion into two modes (\figref{fig:experiment_vortex}(a)) with the same guidelines as \sref{section_pick_place}: excavation at the highest point of the soil pile based on the pixel values in the depth image in manual mode ($o_t=0$), and dumping the excavated soil in the front hole in the automatic mode ($o_t=1$). The mode transition is limited to $o_t:0\rightarrow1\rightarrow0$ in the same manner as \sref{section_pick_place}.

\subsubsection{Demonstration setting} 

We defined the states and the actions as follows:
\begin{itemize*}
\item[\textbf{(i) action policy w/o PA:}] the states are 4-DOF joint angles and velocities, the height of the soil pile and its position $(X, Y)$ in the depth image, and the number of excavations. The actions are the bucket velocities and joint angular velocity $(\dot{X},\dot{Y},\dot{Z},\dot{\theta})$;
\item [\textbf{(ii) action policy with PA:}] the states and actions are identical as in (i); 
\item [\textbf{(iii) mode-switching policy with PA:}] the states are identical as in (i), and the action is the mode. 
\end{itemize*} 
Automatic action policy $\pi^{\mathrm{Auto}}(o_t)$ is defined as follows: if $o_t = 1$, moving the bucket forward while opening its bucket joint \taharamod{$(\dot{X},\dot{Y},\dot{Z},\dot{\theta}) = (0.6~\mathrm{m/s}, 0.0~\mathrm{m/s}, 1.0~\mathrm{m/s}, -0.1~\mathrm{rad/s})$}.

\subsubsection{Learning setting}

The setting is the same as\sref{section_pick_place}. Each episode has about 400 steps.

\subsubsection{Results}

\figref{fig:experiment_vortex}(b) shows the iterative test performances as a more detailed analysis (unlike \figref{fig:experiment_pick_place}(b)). DIPA performed excavations robustly with a faster performance convergence than DIPA(-) and the other comparisons. BC and BCPA showed a significant performance degradation due to a lack of robustness against modeling errors to learn actions that contain high uncertainties in the interaction with the soil. \taharamod{DART also fails the task due to significant modeling errors against the long-horizon task. This result simultaneously caused the problem in the disturbance's level updating in \figref{fig:experiment_vortex}(c). DART maintains a high disturbance's level for all the iterations, which is undesirable for its performance for the same reason as \sref{section_pick_place}.} In contrast, DIPA shows the convergence of the disturbance's level in a few iterations. This result is the key feature of our disturbance optimization on mode-predictive covariate shift in \equref{eq:semi-automatic_disturbance_update}. If the mode-switching policy selects the automatic action policy as a fatal false prediction, this large error is reflected in the disturbance's level based on the difference between the predicted and demonstrated actions. Importantly, injecting this optimized disturbance contributes to the rapid performance improvement at the second iteration (\figref{fig:experiment_vortex}(b)). On the other hand, DIPA(-) immediately shows a low disturbance's level, and its performance convergence is slow due to the lack of robustness of the mode-switching policy.

\section{User study with a real robot}
\label{section_real_robot}

This section evaluates our method for a long-horizon soil-filling task. This validation aims to confirm the test performance and demonstration burden using a user survey similar to a previous work \cite{hoque2021thriftydagger} for the comparisons shown in \tableref{table:ablations} where BC that had the lowest performance and S-DIPA(-) are eliminated. \taharamod{We employed three human subjects with basic knowledge of robot control functioned as operators to confirm the demonstration burden. Unlike simulations, it is challenging to design an algorithmic controller by hand based on this task's states, such as force and image features.}

\subsubsection{Task setting}
\figref{fig:experiment_real_robot}(a) shows the real robot environment. This experiment's purpose is a proof of concept; plastic beads with low shear strength and high flow velocity were used as artificial soil instead of actual soil. The task goal is to fill the dump area with soil, and the success is evaluated by the occupancy ratio in the processed image of the dump area (success exceeds 85~\%). The robot arm (Universal Robot UR5e) has a bucket to excavate the soil. We used a force-based compliance controller to avoid damage when the bucket touches hard objects. A depth camera (Intel RealSense D455) was attached to the environment to capture the soil shape, and the image is updated when there is no occlusion. The operator controlled the robot using a haptic controller (3D SYSTEMS Touch) with force feedback and changed the modes by pushing the button on the device.

\subsubsection{Partial Automation setting}

We decomposed the robot motion into the four modes shown in \figref{fig:experiment_real_robot}(a): excavating the soil pile with force-feedback in the manual mode ($o_t=0$); moving to the dump area in the automatic mode ($o_t=1$); precisely dropping the soil in the dump area to fill it in the manual mode ($o_t=2$); moving to the soil pile in the automatic mode ($o_t=3$). The mode transition is limited to $o_t:0\rightarrow1\rightarrow2\rightarrow3\rightarrow0$. The subjects are instructed to perform the manual and mode-switching operations under PA, and only DART requires a full-manual operation.

\subsubsection{Demonstration setting}

The states and actions are defined as follows:
\begin{itemize*}
\item[\textbf{(i) action policy w/o PA:}] the states are the bucket pose (6-DOF), forces (3-DOF), and the vectorized depth features (25 dim). The actions are the bucket's velocities and the angular velocity $(\dot{X},\dot{Y},\dot{Z},\dot{\theta})$; 
\item[\textbf{(ii) action policy with PA:}] the states and actions are identical as in (i); 
\item[\textbf{(iii) mode-switching policy with PA:}] the states are identical as in (i), and the action is the mode. 
\end{itemize*}
Preliminary experiments confirmed that the force and depth features are necessary since the learned policy was unable to accomplish the task in all five tests without those features in states. 
The automatic action policies $\pi^{\mathrm{Auto}}(o_t)$ are defined as follows: if $o_t = 1$, moving the bucket to the left \taharamod{$(\dot{X},\dot{Y},\dot{Z},\dot{\theta}) = (0.5~\mathrm{m/s}, 0.0~\mathrm{m/s}, 0.0~\mathrm{m/s}, 0.0~\mathrm{rad/s})$}, and if $o_t = 3$, moving the bucket to the right \taharamod{$(\dot{X},\dot{Y},\dot{Z},\dot{\theta}) = (-0.5~\mathrm{m/s}, 0.0~\mathrm{m/s}, 0.0~\mathrm{m/s}, 0.0~\mathrm{rad/s})$.}

\subsubsection{Learning setting}

We collected demonstrations for five iterations with three episodes each (max. of 400 steps per episode), and the learned policies were evaluated five times.

\begin{table}[tb]
\caption{
Demonstration and test results with three human subjects: First and second lines show user survey results (NASA TLX \cite{hart2006nasa}): 1) very low, 2) low, 3) medium, 4) high, and 5) very high. For the mental demand, a significant difference by t-test ($p < 0.05$) was observed between DIPA and DART. For  frustration, a significant difference by t-test ($p < 0.05$) was observed between BCPA and DART.
}
\label{table:result_real_robot}
\centering
\begin{tabular}{c|c c c c}
\hline
  & \textbf{DIPA} & DIPA(-) & BCPA & DART\\
\hline
Mental demand       & \textbf{2.3 $\pm$ 0.5}    & 2.7 $\pm$ 0.9         & 2.3 $\pm$ 1.9             & 4.3 $\pm$ 0.5 \\
Frustration         & 2.7 $\pm$ 0.5             & 3.0 $\pm$ 0.0         & \textbf{1.6 $\pm$ 0.9}    & 4.3 $\pm$ 0.9 \\
Manual steps        & 158 $\pm$ 57              & \textbf{143 $\pm$ 21} & 149 $\pm$ 36              & 296 $\pm$ 78 \\
Success rate [\%]   & \textbf{87 $\pm$ 19}      & 40 $\pm$ 33           & 13 $\pm$ 19               & 7 $\pm$ 9 \\
\hline
\end{tabular}
\end{table}

\begin{figure}[tb]
    \centering
    
    \begin{minipage}[b]{0.22\linewidth}
    \centering
    \includegraphics[width=0.8\hsize]{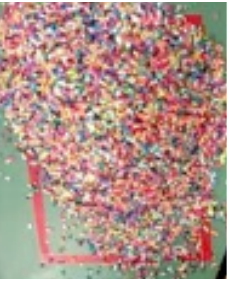}
    \subcaption{DIPA}
    \end{minipage}
    \begin{minipage}[b]{0.22\linewidth}
    \centering
    \includegraphics[width=0.8\hsize]{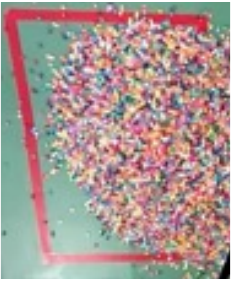}
    \subcaption{DIPA(-)}
    \end{minipage}
    \begin{minipage}[b]{0.22\linewidth}
    \centering
    \includegraphics[width=0.8\hsize]{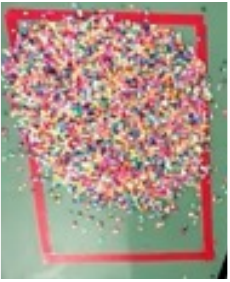}
    \subcaption{BCPA}
    \end{minipage}
    \begin{minipage}[b]{0.22\linewidth}
    \centering
    \includegraphics[width=0.8\hsize]{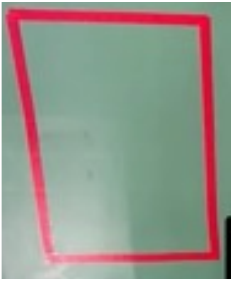}
    \subcaption{DART}
    \end{minipage}
    
\caption{Final scene of dump area in test trial}
\label{fig:result_real_robot_test}
\end{figure}

\subsubsection{Results}

\tableref{table:result_real_robot} shows the demonstration and test results. For the (immediate) mental demand, DIPA, DIPA(-), and BCPA (all of which utilize PA) have smaller burdens than DART, which focuses on learning a single policy. The total manual time steps at the third line also indicate that the methods with PA significantly decrease manual operations. In addition, for the frustration (accumulation of mental load), DIPA, DIPA(-), and BCPA reduced the load more than DART. DIPA and DIPA(-) are more demanding than BCPA due to the disturbance injection, although DIPA and DIPA(-) are not much different. \taharamod{Unlike the significant difference in mental demand and frustration due to the presence of PA or excessive disturbance update in DART, the difference of DIPA and DIPA(-) might be caused by the resolution of the evaluation (from 1 to 5, 1 increment) and the individual differences among the subjects.} These results indicate that utilizing PA significantly reduces the mental load and frustration by resting the collection of the recovery actions from the injected disturbances in each state. This result is also displayed as the disturbance's level in \figref{fig:experiment_real_robot}(b). The last line shows the average test performance of the three subjects, and \figref{fig:result_real_robot_test} shows the final scene of one test trial. Under the human operators' demonstration, our method achieved the best performance compared to the other methods and reduced the demonstration burden.

\section{DISCUSSION}
\taharamod{We proposed a novel imitation learning framework for the tasks with Partial Automation (PA). Our evaluations confirm that our proposed method has the following three advantages: }
\begin{enumerate*}
\item \textbf{sample efficiency}: DIPA outperformed DART under the same amount of training data; 
\item \textbf{robust policies}: DIPA outperformed both DIPA(-) and BCPA owing to the disturbance's level optimization on the mode-predictive covariate shift, and DIPA is also stable under several designs of automatic action policies; 
\item \textbf{less demonstration burden}: DIPA significantly reduced the demonstration burden in the three metrics of the user evaluation compared to DART
\end{enumerate*}.

Our future works will address the following limitations.
\begin{enumerate*}
\item Designing automatic action policies for each mode requires prior knowledge of the task. To alleviate this requirement, unsupervised trajectory segmentation \cite{tsai2019unsupervised}, which extracts segments from demonstrations, could be beneficial.
\item Our method employs human-gated policy switching \cite{Kelly2019HGDAggerII} to utilize flexible human decision-making for demonstrations, even though this might increase the operator's cognitive load. As a feature of interactive IL, robot-gated mode-switching\cite{hoque2021thriftydagger} is an attractive alternative for reducing cognitive load.
\item \taharamod{Although this study focused on PA to the Disturbance Injection approach, it might be a beneficial direction to compare it with other IL approaches that also aim to reduce the covariate shift in different ways, such as DAgger family\cite{ross2011reduction, Kelly2019HGDAggerII, hoque2021thriftydagger} in our PA framework.}
\end{enumerate*}

\section{CONCLUSION}

This paper proposed a novel robust imitation learning framework with disturbance injection under Partial Automation for long-horizon robot tasks. We verified the effectiveness of our method in simulations and a real environment and achieved a better performance than the previous methods while simultaneously reducing the demonstration burden.

\bibliographystyle{IEEEtran}
\bibliography{reference}

\end{document}